\documentclass[letterpaper, 10 pt, journal, twoside]{ieeetran}

\IEEEoverridecommandlockouts

\usepackage{amsmath} %
\usepackage{amssymb}  %
\usepackage[protrusion=false]{microtype}
\usepackage{tensor}
\usepackage{mathtools}
\usepackage{makecell}
\usepackage{todonotes}
\usepackage{siunitx}
\usepackage{booktabs}

\title{Unified Multi-Modal Landmark Tracking \\for Tightly Coupled 
Lidar-Visual-Inertial Odometry}

\author{David Wisth$^1$, Marco Camurri$^1$, Sandipan Das$^{2,3}$, Maurice
Fallon$^1$%
\thanks{Manuscript received: October, 15, 2020; Revised January, 8, 2021;
		Accepted January, 13, 2021.} %
\thanks{This paper was recommended for publication by Editor Eric Marchand
upon evaluation of the Associate Editor and Reviewers' comments. 
This research has been conducted as part of the ANYmal research community. It
was part funded by the EU H2020 Projects THING and MEMMO, a Royal Society
University Research Fellowship (Fallon) and a Google DeepMind studentship
(Wisth). Special thanks to the CERBERUS DARPA SubT Team for providing data
from challenge runs.} %
\thanks{$^1$Oxford Robotics Institute, Department of Engineering Science, 
University of Oxford, UK {\tt\small \{davidw, mcamurri, 
mfallon\}@robots.ox.ac.uk}} %
\thanks{$^2$KTH Royal Institute of Technology, Sweden
{\tt\small sandipan@kth.se}} %
\thanks{$^3$Scania AB, Sweden {\tt\small sandipan.das@scania.com}} %
\thanks{Digital Object Identifier (DOI): see top of this page.}
}

\usepackage[linesnumbered,ruled,vlined]{algorithm2e}

\usepackage{multirow}

\SetCommentSty{sandy}
\SetKwInput{KwInput}{Input} %
\SetKwInput{KwOutput}{Output} %

\usepackage{xcolor} %

\newcommand{\X}{\mathcal{X}}
\newcommand{\Z}{\mathcal{Z}}
\newcommand{\etalcite}[2]{#1~et~al.~\cite{#2}}

\DeclareMathOperator*{\argmax}{arg\,max}
\DeclareMathOperator*{\argmin}{arg\,min}

\newcommand{\hide}[1]{}
\newcommand{\Figure}{Fig.~}
\newcommand{\Equation}{Eq.~}
\newcommand{\ie}{{i.e.,~}}

\newcommand{\bdmath}{\begin{dmath}}
\newcommand{\edmath}{\end{dmath}}
\newcommand{\beq}{\begin{equation}}
\newcommand{\eeq}{\end{equation}}
\newcommand{\bdm}{\begin{displaymath}}
\newcommand{\edm}{\end{displaymath}}
\newcommand{\bea}{\begin{eqnarray}}
\newcommand{\eea}{\end{eqnarray}}
\newcommand{\beal}{\beq \begin{array}{ll}}
\newcommand{\eeal}{\end{array} \eeq}
\newcommand{\beas}{\begin{eqnarray*}}
\newcommand{\eeas}{\end{eqnarray*}}
\newcommand{\ba}{\begin{array}}
\newcommand{\ea}{\end{array}}
\newcommand{\bit}{\begin{itemize}}
\newcommand{\eit}{\end{itemize}}
\newcommand{\ben}{\begin{enumerate}}
\newcommand{\een}{\end{enumerate}}

\newcommand{\SO}{\mathrm{SO}}

\newcommand{\Real}{\mathbb{R}}

\newcommand{\SEthree}{\ensuremath{\mathrm{SE}(3)}\xspace}

\newcommand{\SOthree}{\ensuremath{\SO(3)}\xspace}

\newcommand{\calC}{{\cal C}}

\newcommand{\calI}{{\cal I}}

\newcommand{\calL}{{\cal L}}

\newcommand{\calX}{{\cal X}}
\newcommand{\calZ}{{\cal Z}}

\newcommand{\T}{\mathbf{T}}
\newcommand{\R}{\mathbf{R}}

\newcommand{\transpose}{\mathsf{T}}

\newcommand{\rotvel}{\boldsymbol\omega}

\newcommand{\tran}{\mathbf{p}}

\newcommand{\vel}{\mathbf{v}}

\newcommand{\bias}{\mathbf{b}}

\newcommand{\logmap}{\mathrm{Log}}

\newcommand{\State}{\boldsymbol{x}}

\newcommand{\World}{\mathtt{W}}
\newcommand{\Imu}{\mathtt{I}}
\newcommand{\Camera}{\mathtt{C}}
\newcommand{\Lidar}{\mathtt{L}}
\newcommand{\world}{\mathtt{{W}}}

\newcommand{\imu}{\mathtt{{I}}}
\newcommand{\base}{\mathtt{{B}}}

\newcommand{\Base}{\mathtt{{B}}}

\newcommand{\meters}{\rm{m}}

\newcommand{\Landmark}{\boldsymbol{m}}
\newcommand{\Plane}{\boldsymbol{p}}

\newcommand{\Line}{\boldsymbol{l}}

\newcommand{\Log}[1]{\logmap \left( #1 \right)}

\newcommand{\defeq}{\triangleq}

\makeatletter
\let\NAT@parse\undefined
\makeatother
\usepackage{hyperref}
\begin{document}

\onecolumn
\thispagestyle{empty}

\hspace{3cm}
\begin{center}
This paper has been accepted for publication in \emph{IEEE Robotics And Automation Letters} (RA-L).\\

\hspace{1cm}

DOI: \href{http://dx.doi.org/10.1109/LRA.2021.3056380}{10.1109/LRA.2021.3056380}\\

IEEE Explore: \url{https://ieeexplore.ieee.org/document/9345356} \\

\hspace{1cm}

Please cite the paper as: \\

\hspace{1cm}

D. Wisth, M. Camurri, S. Das and M. Fallon, \\
``Unified Multi-Modal Landmark  Tracking for Tightly Coupled
Lidar-Visual-Inertial Odometry,'' \\
in \emph{IEEE Robotics and Automation Letters},
vol. 6, no. 2, pp. 1004-1011, April 2021 \\
\end{center}
\twocolumn

\setcounter{page}{1}
\maketitle

\begin{abstract}
We present an efficient multi-sensor odometry system for mobile platforms that
jointly optimizes visual, lidar, and inertial information within a single
integrated factor graph. This runs in real-time at full framerate using fixed
lag smoothing. To perform such tight integration, a new method to extract 3D
line and planar primitives from lidar point clouds is presented. This approach
overcomes the suboptimality of typical frame-to-frame tracking methods by
treating the primitives as landmarks and tracking them over multiple scans. True
integration of lidar features with standard visual features and IMU is made
possible using a subtle passive synchronization of lidar and camera frames. The
lightweight formulation of the 3D features allows for real-time execution on a
single CPU. Our proposed system has been tested on a variety of platforms and
scenarios, including underground exploration with a legged robot and outdoor
scanning with a dynamically moving handheld device, for a total duration of
\SI{96}{\minute} and \SI{2.4}{\kilo\meter} traveled distance. In these test
sequences, using only one exteroceptive sensor leads to failure due to either
underconstrained geometry (affecting lidar) or textureless areas caused by
aggressive lighting changes (affecting vision). In these conditions, our factor
graph naturally uses the best information available from each sensor modality
without any hard switches.
\end{abstract}


\section{Introduction}
\label{sec:introduction}
\begin{figure}
 \centering
 \includegraphics[width=1\columnwidth]{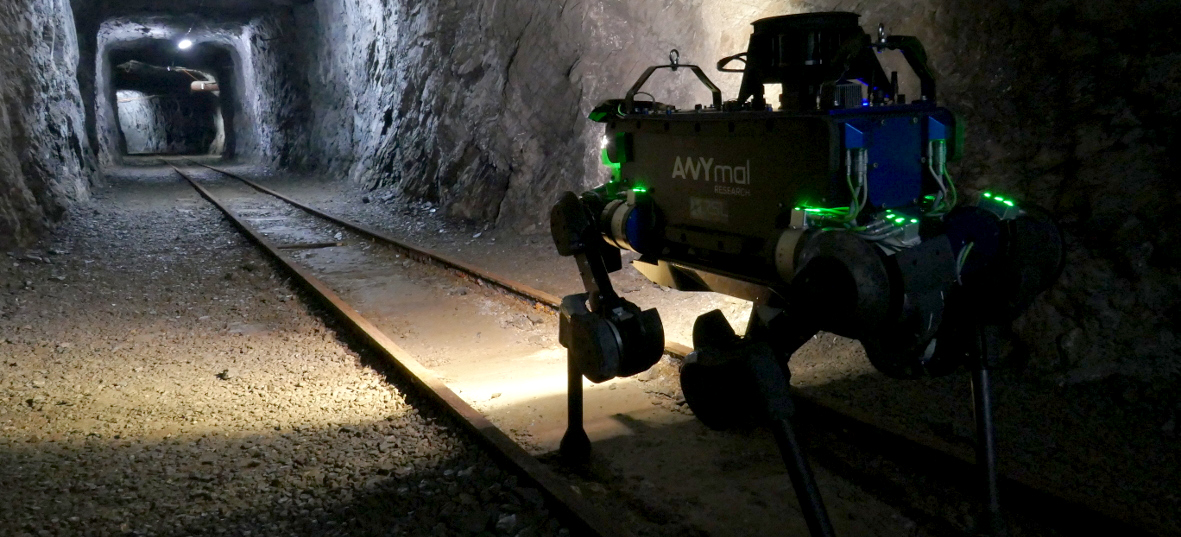}\\
 \vspace{1mm}
 \includegraphics[width=1\columnwidth]{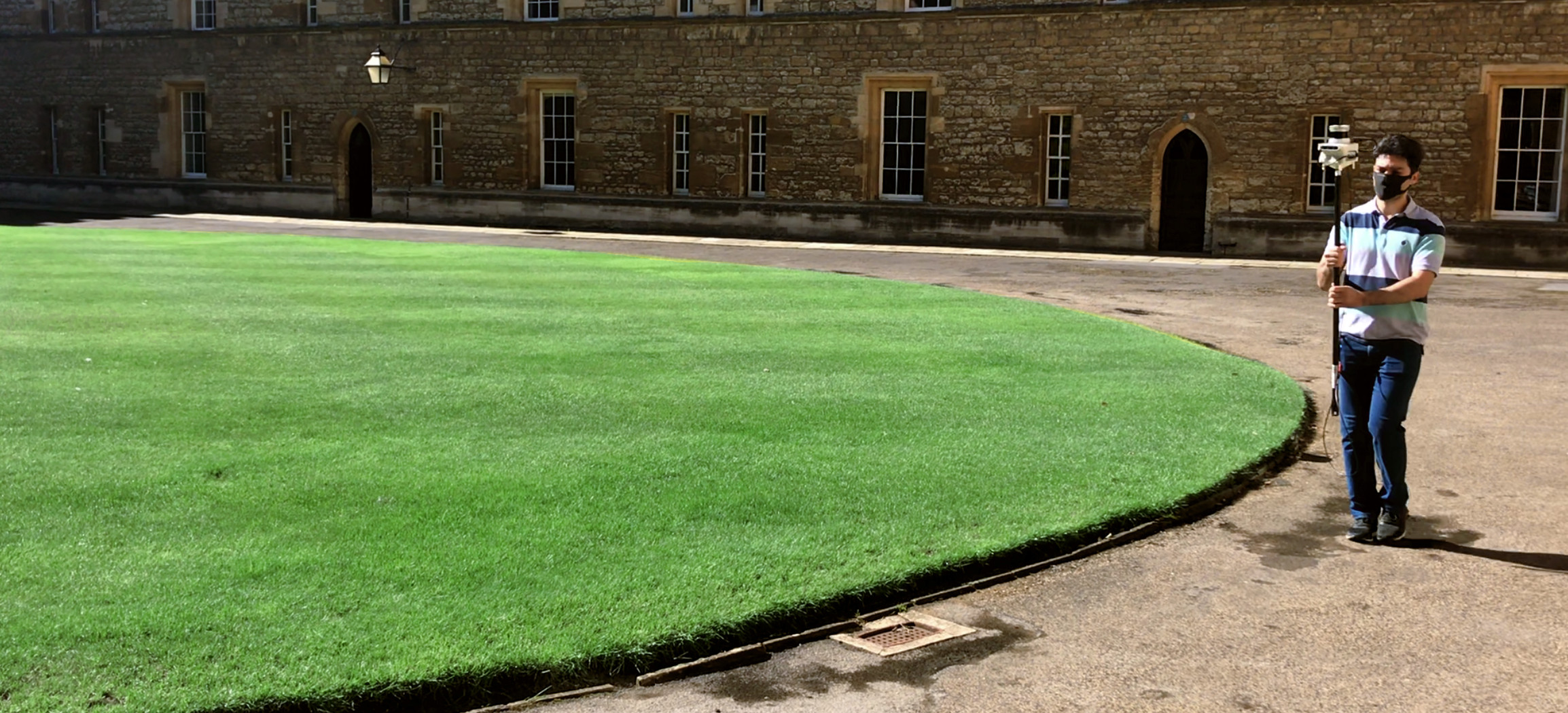}
 \caption{We tested our multi-sensor odometry algorithm with data from the 
 ANYmal quadruped robot \cite{Hutter2016} in the 
 DARPA SubT Challenge (top, courtesy RSL/ETH Zurich) and
 a handheld mapping device in New College, Oxford
\cite{ramezani2020newer} (bottom). Video: 
\texttt{\url{https://youtu.be/MjXYAHurWe8}}}
 \label{fig:anymal-mine}
 \vspace{-2mm}
\end{figure}

\IEEEPARstart{M}{ulti-modal} sensor fusion is a key capability required for
successful deployment of autonomous navigation systems in real-world scenarios.
The spectrum of possible applications is wide, ranging from autonomous
underground exploration  to outdoor dynamic mapping (see \Figure
\ref{fig:anymal-mine}).

Recent advances in lidar hardware have fostered research into lidar-inertial
fusion \cite{Zhang_2014,shan-legoloam,lips,ye2019icra}. The wide field-of-view,
density, range, and accuracy of lidar sensors makes them suitable for
navigation, localization, and mapping tasks.

However, in environments with degenerate geometries, such as long tunnels or
wide open spaces, lidar-only approaches can fail. Since the IMU integration
alone cannot provide reliable pose estimates for more than a few seconds, the
system failure is often unrecoverable. To cope with these situations, fusion
with additional sensors, in particular cameras, is also required. While
visual-inertial-lidar fusion has already been achieved in the past via loosely
coupled methods \cite{pronto}, tightly coupled methods such as incremental
smoothing are more desirable because of their superior robustness.

In the domain of smoothing methods, research on Visual-Inertial Navigation
Systems (VINS) is now mature and lidar-inertial systems are becoming
increasingly popular. However, the tight fusion of all three sensor modalities
at once is still an open research problem.

\subsection{Motivation}
The two major challenges associated with the fusion of IMU, lidar and camera
sensing are: 1) achieving real-time performance given the limited computational
budget of mobile platforms and 2) the appropriate synchronization of three
signals running at different frequencies and methods of acquisition.

Prior works have addressed these two problems by adopting loosely coupled
approaches \cite{pronto, vloam, wang2019iros}  or by running two separate
systems (one for lidar-inertial and the other for visual-inertial odometry)
\cite{shao2019iros}.

Instead, we are motivated to tackle these problems by: 1) extracting and
tracking sparse lightweight primitives and 2) developing a coherent factor graph
which leverages IMU pre-integration to transform dynamically dewarped point
clouds to the timestamp of nearby camera frames. The former avoids matching
entire point clouds (e.g. ICP) or tracking hundreds of feature points (as in
LOAM \cite{Zhang_2014}). The latter makes real-time smoothing of all the sensors
possible.

\subsection{Contribution}
\label{sec:contribution}

The main contributions of this work are the following:
\begin{itemize}
	\item A novel factor graph formulation that tightly fuses vision, lidar and
IMU measurements within a single consistent optimization process;
	\item An efficient method for extracting lidar features, which are then
optimized as landmarks. Both lidar and visual features share a unified
representation, as the landmarks are all treated as $n$-dimensional parametric
manifolds (\ie points, lines and planes). This compact representation allows us
to process all the lidar scans at nominal framerate;
    \item Extensive experimental evaluation across a range of scenarios
demonstrating superior robustness when compared to more typical approaches which
struggle when individual sensor modalities fail.
\end{itemize}

Our work builds upon the VILENS estimation system introduced in our previous
works \cite{Wisth2019,wisth2020icra} by adding lidar feature tracking and
lidar-aided visual tracking. The combination of camera and lidar enables the use
on portable devices even when moving aggressively, as it naturally handles
degeneracy in the scene (either due to a lack of lidar or visual features).

\section{Related Work}
\label{sec:related-work}
Prior works on multi-modal sensor fusion use combinations of lidar, camera and
IMU sensing and can be characterised as either loosely or tightly coupled, as
summarized in Table \ref{tab:odom_classification}. Loosely coupled systems
process the measurements from each sensor separately and fuse them within a
filter, where they are marginalized to get the current state estimate.
Alternatively, tightly coupled systems jointly optimize both past and current
measurements to obtain a complete trajectory estimate.

Another important distinction is between odometry systems and SLAM systems. In
the latter, loop closures are performed to keep global consistency of the
estimate once the same place is visited twice. Even though some of the works in
the table also include a pose-graph SLAM backend, we are mainly interested in
high-frequency odometry here.

\begin{table}[!htbp]
\normalsize
\centering
\resizebox{\columnwidth}{!}{
\begin{tabular}{p{0.15\linewidth}p{0.52\linewidth}p{0.43\linewidth}}
\toprule
\textbf{Sensors} & \textbf{Loosely coupled systems} & \textbf{Tightly coupled
systems}\\
\midrule
\multirow{3}{*}{\makecell[l]{Lidar + \\ IMU}} &
\multirow{3}{*}{\makecell[l]{LOAM$^{\star}$\cite{Zhang_2014},
LEGO-LOAM$^{\star}$\cite{shan-legoloam}, \\
SLAM with point and \\
plane\textsuperscript{\dag}\cite{grant2018auro}}} &
\multirow{3}{*}{\makecell[l]{LIPS\cite{lips}, LIOM\cite{ye2019icra},\\
LIO-SAM\cite{tixiao2020lio-sam}}}\\
& & \\
& & \\
\midrule %
\multirow{4}{*}{\makecell[l]{Vision + \\ Lidar + \\ IMU}} &
\multirow{4}{*}{\makecell[l]{V-LOAM\cite{vloam}, Visual-Inertial-\\Laser
SLAM\cite{wang2019iros}, Multi-modal \\Sensor
Fusion\cite{Khattak2020}\textsuperscript{\ddag},
Pronto\cite{pronto}}} & \multirow{4}{*}{\makecell[l]{LIMO\cite{limo},
VIL-SLAM\cite{shao2019iros}, \\
INS with Point and Plane \\feature\cite{yang2019icra},
LIC-Fusion\cite{lic-fusion}, \\LIC-Fusion 2.0\cite{zuo2020licfusion}}} \\
& & \\
& & \\
& & \\
\bottomrule
\multicolumn{3}{l}{\textit{$^{\star}$ IMU is optional},
\textit{\textsuperscript{\dag} Lidar only solution},
\textit{\textsuperscript{\ddag}  Additional thermal camera.}} \\
\end{tabular}}
\caption{Different multi-modal odometry estimation algorithms}
\label{tab:odom_classification}
\vspace{-6mm}
\end{table}

\subsection{Loosely Coupled Lidar-Inertial Odometry}
Lidar-based odometry has gained popularity thanks to the initial work of
\etalcite{Zhang}{Zhang_2014}, who proposed the LOAM algorithm. One of their key
contributions is the definition of edge and planar 3D feature points which are
tracked frame-to-frame. The motion between two frames is linearly interpolated
using an IMU running at high-frequency. This motion prior is used in the fine
matching and registration of the features to achieve high accuracy odometry.
\etalcite{Shan}{shan-legoloam} proposed LeGO-LOAM, which further improved the
real-time performance of LOAM for ground vehicles by optimizing an estimate of
the ground plane.

However, these algorithms will struggle to perform robustly in structure-less
environments or in degenerate scenarios \cite{Zhang2016} where constraints
cannot be found due to the lidar's limited range and resolution --- such as long
motorways, tunnels, and open spaces.

\subsection{Loosely Coupled Visual-Inertial-Lidar Odometry}
In many of the recent works \cite{vloam, wang2019iros, Khattak2020, pronto} 
vision was incorporated along with lidar and IMU for odometry estimation in a
loosely coupled manner to provide a complementary sensor modality to both avoid
degeneracy and have a smoother estimated trajectory over lidar-inertial systems.

The authors of LOAM extended their algorithm by integrating feature tracking
from a monocular camera in V-LOAM \cite{vloam} along with IMU, thereby
generating a visual-inertial odometry prior for lidar scan matching. However,
the operation was still performed frame-to-frame and didn't maintain global
consistency. To improve consistency, a Visual-Inertial-Lidar SLAM system was
introduced by \etalcite{Wang}{wang2019iros} where they used a V-LOAM based
approach for odometry estimation and performed a global pose graph optimization
by maintaining a keyframe database. \etalcite{Khattak}{Khattak2020} proposed
another loosely coupled approach similar to V-LOAM, that uses a visual/thermal
inertial prior for lidar scan matching. To overcome degeneracy, the authors used
visual and thermal inertial odometry so as to operate in long tunnels with no
lighting. In Pronto \cite{pronto}, the authors used visual-inertial-legged
odometry as a motion prior for a lidar odometry system and integrated pose
corrections from visual and lidar odometry to correct pose drift in a loosely
coupled manner.

\subsection{Tightly Coupled Inertial-Lidar Odometry}
One of the earlier methods to tightly fuse lidar and IMU was proposed in LIPS
\cite{lips}, a graph-based optimization framework which optimizes the 3D plane
factor derived from the closest point-to-plane representation along with
preintegrated IMU measurements. In a similar fashion, \etalcite{Ye}{ye2019icra}
proposed LIOM, a method to jointly minimize the cost derived from lidar features
and pre-integrated IMU measurements. This resulted in better odometry estimates
than LOAM in faster moving scenarios. \etalcite{Shan}{tixiao2020lio-sam}
proposed LIO-SAM, which adapted the LOAM framework by introducing scan matching
at a local scale instead of global scale. This allowed new keyframes to be
registered to a sliding window of prior ``sub-keyframes'' merged into a voxel
map. The system was extensively tested on a handheld device, ground, and
floating vehicles, highlighting the quality of the reconstruction of the SLAM
system. For long duration navigation they also used loop-closure and GPS factors
for eliminating drift.

Again, due to the absence of vision, the above algorithms may struggle to
perform robustly in degenerate scenarios.

\subsection{Tightly Coupled Visual-Inertial-Lidar Odometry}
To avoid degeneracy and to make the system more robust, tight integration of
multi-modal sensing capabilities (vision, lidar, and IMU) was explored in some
more recent works \cite{shao2019iros, limo, yang2019icra, lic-fusion,
zuo2020licfusion}. In LIMO \cite{limo} the authors presented a bundle
adjustment--based visual odometry system. They combined the depth from lidar
measurements by re-projecting them to image space and associating them to the
visual features which helped to maintain accurate scale.
\etalcite{Shao}{shao2019iros} introduced VIL-SLAM where they combined VIO along
with lidar-based odometry as separate sub-systems for combining the different
sensor modalities rather than doing a joint optimization.

To perform joint state optimization, many approaches \cite{yang2019icra,
lic-fusion, zuo2020licfusion} use the Multi-State Constraint Kalman Filter
(MSCKF) framework \cite{msckf}. \etalcite{Yang}{yang2019icra} tightly integrated
the plane features from an RGB-D sensor within \SI{3.5}{\meter} range and point
features from vision and IMU measurements using an MSCKF. To limit the state
vector size, most of the point features were treated as MSCKF features and
linearly marginalized, while only a few point features enforcing point-on-plane
constraints were kept in state vector as SLAM features. LIC-Fusion introduced by
\etalcite{Zuo}{lic-fusion} tightly combines the IMU measurements, extracted
lidar edge features, as well as sparse visual features, using the MSCKF fusion
framework. Whereas, in a recent follow up work, LIC-Fusion 2.0
\cite{zuo2020licfusion}, the authors introduced a sliding window based
plane-feature tracking approach for efficiently processing 3D lidar point
clouds.

In contrast with previous works, we jointly optimize the three aforementioned
sensor modalities within a single, consistent factor graph optimization
framework. To process lidar data at real-time, we directly extract and track 3D
primitives such as lines and planes from the lidar point clouds, rather than
performing  ``point-to-plane'' or ``point-to-line'' based cost functions. This
allows for natural tracking over multiple frames in a similar fashion to visual
tracking, and to constrain the motion even in degenerate scenarios.

\section{Problem Statement}
\label{sec:problem-statement}

We aim to estimate the position, orientation, and linear velocity of a mobile
platform (in our experiments, a legged robot or a handheld sensor payload)
equipped with IMU, lidar and either a mono or stereo camera with low latency and
at full sensor rate.

\begin{figure}
\centering
\vspace{2mm}
\includegraphics[width=\columnwidth]{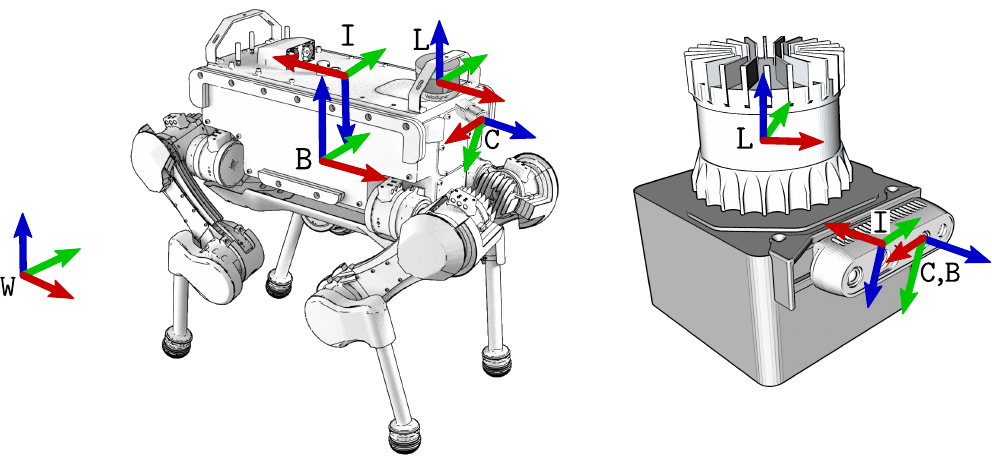}
\caption{Reference frames conventions for the ANYmal platform and the handheld
device. The world frame $\World$ is a fixed frame, while the base frame $\Base$,
camera optical frame $\Camera$, IMU frame, $\Imu$, and lidar frame $\Lidar$ are
attached to the moving robot's chassis or device. For simplicity, $\Camera$ and
$\Base$ are coincident on the handheld device. }
\label{fig:coordinate-frames}
\vspace{-5mm}
\end{figure}

The relevant reference frames are specified in \Figure
\ref{fig:coordinate-frames} and include the robot-fixed base frame $\Base$, left
camera frame $\Camera$, IMU frame $\Imu$, and lidar frame $\Lidar$. We wish to
estimate the position of the base frame relative to a fixed world frame
$\World$.

Unless otherwise specified, the position
$\tensor[_\world]{\tran}{_{\world\base}}$ and orientation $\R_{\world\base}$ of
the base (with $\tensor[_\world]{\T}{_{\world\base}} \in \SEthree$ as the
corresponding homogeneous transform) are expressed in world coordinates;
velocities $\tensor[_\base]{\vel}{_{\world\base}},
\tensor[_\base]{\rotvel}{_{\world\base}}$ are in the base frame, and IMU biases
$\tensor[_\imu]{\bias}{^{g}},\;\tensor[_\imu]{\bias}{^{a}}$ are expressed in the
IMU frame.

\subsection{State Definition}
\noindent The mobile platform state at time $t_i$ is defined as follows:
\begin{equation} \State_i \triangleq \left[\R_i,\tran_i,\vel_i, \bias^{g}_i \,\;
\bias^{a}_i \right] \in \SOthree \times \Real^{12} \end{equation} where: $\R_i$
is the orientation, $\tran_i$ is the position, $\vel_i$ is the linear velocity,
and the last two elements are the usual IMU gyroscope and accelerometer biases.

In addition to the states, we track the parameters of three $n$-manifolds:
points, lines and planes. The point landmarks $\Landmark_{\ell}$ are visual
features, while lines $\Line_{\ell}$ and planes $\Plane_{\ell}$ landmarks are
extracted from lidar. The objective of our estimation are all states and
landmarks visible up to the current time $t_k$:

\begin{equation}
\X_k \defeq \left[
\bigcup_{\forall i \in \mathsf{X}_k} \lbrace \State_{i} \rbrace,
\bigcup_{\forall \ell \in \mathsf{M}_k} \lbrace \Landmark_{\ell} \rbrace,
\bigcup_{\forall \ell \in \mathsf{P}_k} \lbrace \Plane_{\ell} \rbrace,
\bigcup_{\forall \ell \in \mathsf{L}_k} \lbrace \Line_{\ell} \rbrace
\right]
\end{equation}
where $\mathsf{X}_k, \mathsf{M}_k, \mathsf{P}_k, \mathsf{L}_k$ are the lists of
all states and landmarks tracked within a fixed lag smoothing window.

\subsection{Measurements Definition}
The measurements from a mono or stereo camera $\calC$, IMU $\calI$, and lidar
$\calL$ are received at different times and frequencies. We define $\Z_k$ as the
full set of measurements received within the smoothing window. Subsection
\ref{sec:lidar-sync} explains how the measurements are integrated within the
factor graph, such that the optimization is performed at fixed frequency.

\subsection{Maximum-a-Posteriori Estimation}
We maximize the likelihood  of the measurements, $\calZ_k$, given the history of
states, $\calX_k$: \begin{equation} \X^*_k = \argmax_{\X_k} p(\X_k|\Z_k) \propto
p(\X_0)p(\Z_k|\X_k) \label{eq:posterior} \end{equation} The measurements are
formulated as conditionally independent and corrupted by white Gaussian noise.
Therefore, \Equation (\ref{eq:posterior}) can be formulated as the following
least squares minimization problem \cite{Dellaert2017}:
\begin{multline}
\X^{*}_k = \argmin_{\X_k}
\sum_{i \in \mathsf{K}_k} \left(
\|\mathbf{r}_{\calI_{ij}}\|^2_{\Sigma_{\calI_{ij}}}
+ \sum_{\ell \in \mathsf{P}_i} \| \mathbf{r}_{\State_i, \Plane_\ell}
\|^2_{\Sigma_{\State_i, \Plane_\ell} } +
\right. \\
\left. \sum_{\ell \in \mathsf{L}_i} \| \mathbf{r}_{\State_i, \Line_\ell}
\|^2_{\Sigma_{\State_i, \Line_\ell} }
+ \sum_{\ell \in \mathsf{M}_i} \|\mathbf{r}_{\State_i,\Landmark_{\ell}}
\|^2_{\Sigma_{\State_i, \Landmark_\ell}}
\right) + \|\mathbf{r}_0\|^2_{\Sigma_0}
\end{multline}
where $\calI_{ij}$ are the IMU measurements between $t_{i}$ and $t_j$ and
$\mathsf{K}_k$ are all keyframe indices before $t_k$. Each term is the residual
associated to a factor type, weighted by the inverse of its covariance matrix.
The residuals are: IMU, lidar plane and line features, visual landmarks, and a
state prior.

\begin{figure}
 \centering
 \includegraphics[width=1.0\columnwidth]{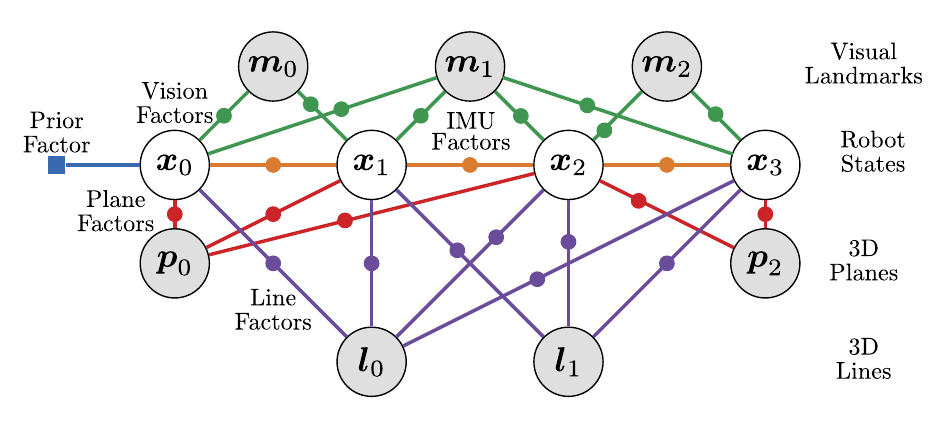}
 \caption{VILENS sliding-window factor graph structure, showing prior, visual,
plane, line, and preintegrated IMU factors. Tracking landmarks over time
increases the accuracy of the estimation by allowing new measurements to improve
the accuracy of past states.}
\label{fig:factor-graph}
\vspace{-3mm}
\end{figure}

\section{Factor Graph Formulation}
\label{sec:twist-factors}

We now describe the measurements, residuals, and covariances of the factors in
the graph, shown in \Figure \ref{fig:factor-graph}. For convenience, we
summarize the IMU factors in Section \ref{sec:imu-factors}; then, we introduce
the visual-lidar landmark factors in Sections \ref{sec:visual-factors} and
\ref{sec:stereo-factors}, while Section \ref{sec:plane-line-factors} describes
our novel plane and line landmark factors.

\subsection{Preintegrated IMU Factors}
\label{sec:imu-factors}
We follow the now standard manner of IMU measurement preintegration
\cite{Forster2017} to constrain the pose, velocity, and biases between two
consecutive nodes of the graph, and provide high frequency states updates
between nodes. The residual has the form:
\begin{equation}
\mathbf{r}_{\calI_{ij}}  = \left[ \mathbf{r}^\transpose_{\Delta
\R_{ij}}, \mathbf{r}^\transpose_{\Delta \vel_{ij}},
\mathbf{r}^\transpose_{\Delta \tran_{ij}},
\mathbf{r}_{\bias^a_{ij}},
\mathbf{r}_{\bias^g_{ij}} \right]
\end{equation}
For the definition of the residuals, see \cite{Forster2017}.

\subsection{Mono Landmark Factors with Lidar Depth}
\label{sec:visual-factors}

To take full advantage of the fusion of vision and lidar sensing modalities, we
track monocular visual features but use the lidar's overlapping field-of-view to
provide depth estimates, as in \cite{limo}. To match the measurements from lidar
and camera, which operate at 10 Hz and 30 Hz respectively, we use the method
described in Section \ref{sec:lidar-sync}.

\begin{figure}
 \vspace{2mm}
 \centering
 \includegraphics[height=2.47cm]{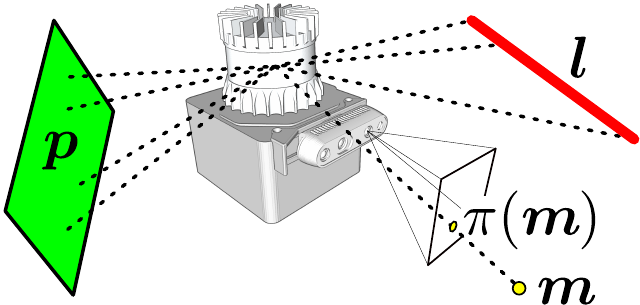}
 \includegraphics[height=2.47cm]{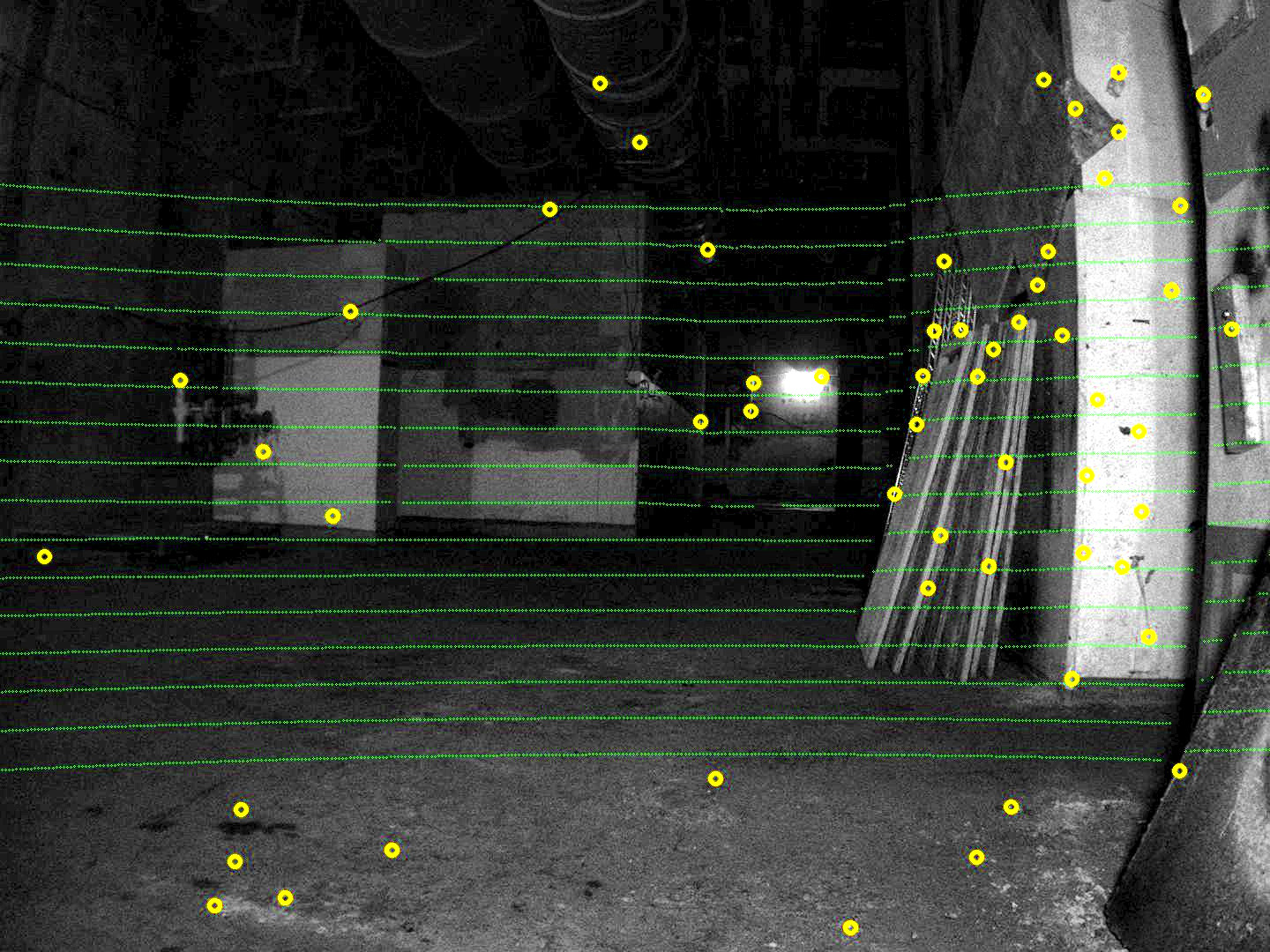}
 \caption{\textit{Left}: The visual FAST features $\Landmark$ (yellow); the
lidar line $\Line$: (red), and plane $\Plane$ primitive (green) are tracked by
our method. \textit{Right}: The projection of the lidar data (green) along with
the visual features (yellow) into the image frame, which helps to associate
depth to the visual features.}
 \label{fig:features}
 \vspace{-3mm}
\end{figure}

Let $\Landmark_\ell \in \Real^3$ be a visual landmark in Euclidean space, $\pi:
\SEthree \times \Real^3 \mapsto \Real^2$ a function that projects a landmark to
the image plane given a platform pose $\T_i$ (for simplicity, we omit the fixed
transforms between base, lidar and camera), and $(u_{\ell}, v_{\ell}) \in
\Real^2$ a detection of $\Landmark_\ell$ on the image plane (yellow dots in
\Figure \ref{fig:features}, right). We first project all the points
$\tilde{\mathbf{x}}_m \in \calL_i$ acquired by the lidar between time $t_i$ and
$t_{i+1}$ onto the image plane with $\pi(\T_i, \tilde{\mathbf{x}}_m)$ (green
dots in \Figure \ref{fig:features}, right). Then, we find the projected point
$\pi(\tilde{\mathbf{x}}_\ell)$ that is closest to $(u_{\ell}, v_{\ell})$ on the
image plane within a neighborhood of 3 pixels. Finally, the residual is computed
as:
\begin{equation}
\mathbf{r}_{\State_i, \Landmark_\ell} = \T^{-1}_i \Landmark_\ell  -
\tilde{\mathbf{x}}_\ell
\end{equation}
When we cannot associate lidar depth to a visual feature (due to the different
resolution of lidar and camera sensors) or if it is unstable (\ie when the depth
changes $> \SI{0.5}{\meter}$ between frames due to dynamic obstacles or noise),
we revert to stereo matching, as described in the next section.

\subsection{Stereo Landmark Factors}
\label{sec:stereo-factors}

The residual at state $\State_i$ for landmark $\Landmark_\ell$ is
\cite{wisth2020icra}:
\begin{equation}
	\mathbf{r}_{\State_i, \Landmark_\ell} =
	\left( \begin{array}{c}
		\pi_u^L(\T_i, \Landmark_\ell) - u^L_{i,\ell} \\
		\pi_u^R(\T_i, \Landmark_\ell) - u^R_{i,\ell} \\
		\pi_v(\T_i, \Landmark_\ell) - v_{i,\ell}
	\end{array} \right) \label{eq:stereo-residual}
\end{equation}
where $(u^{L}, v), (u^{R}, v)$ are the pixel locations of the detected landmark
and $\Sigma_{\Landmark}$ is computed from an uncertainty of \num{0.5} pixels.
Finally, if only a monocular camera is available, then only the first and last
elements in \Equation \ref{eq:stereo-residual} are used.

\subsection{Plane Landmark Factor}
\label{sec:plane-line-factors}

We use the Hessian normal form to parametrize an infinite plane $\Plane$ as a
unit normal $\hat{\mathbf{n}} \in \Real^3$ and a scalar $d$ representing its
distance from the origin:
\begin{equation}
\Plane = \left\{\langle \hat{\mathbf{n}}, d \rangle \in
\mathbb{R}^{4} ~\vert~ \hat{\mathbf{n}}\cdot (x,y,z) + d = 0 \right\}
\end{equation}
Let $\otimes$ be the operator that applies a homogeneous transform $\T$ to all
the points of a plane $\Plane$, and $\ominus$ the operator that defines the
error between two planes $(\Plane_i,\Plane_j)$ as:
\begin{equation}
\Plane_i \ominus \Plane_j = \left( B_p^\transpose \hat{\xi}, d_i - d_j \right)
\in \mathbb{R}^{3}
\label{eq:plane}
\end{equation}
where $B_p \in \Real^{3 \times 2}$ is a basis for the tangent space of
$\hat{\mathbf{n}}_i$ and  $\hat{\xi}$ is defined as follows \cite{dellaertmath}:
\begin{equation}
 \hat{\xi} =
-\frac{\arccos{(\hat{\mathbf{n}}_i\cdot\hat{\mathbf{n}}_j)}}{1-(\hat{\mathbf{n}}
_i\cdot\hat{\mathbf{n}}_j)^2} \left(\hat{\mathbf{n}}_j -
(\hat{\mathbf{n}}_i\cdot\hat{\mathbf{n}}_j)\hat{\mathbf{n}}_i\right) \in 
\Real^3
\end{equation}
When a plane $\tilde{\Plane}_i$ is measured at time $t_i$, the corresponding
residual is the difference between $\tilde{\Plane}$ and the estimated plane
$\Plane_\ell$ transformed into the local reference frame:
\begin{equation}
\mathbf{r}_{x_i, \Plane_\ell} = \left({\T}^{-1}_i \otimes \Plane_\ell \right) 
\ominus
\tilde{\Plane}_i
\label{eq:plane-residual}
\end{equation}

\subsection{Line Landmark Factor}
Using the approach from \cite{minlie}, infinite straight lines  can be
parametrized by a rotation matrix $\R \in \SOthree$ and two scalars $a$, $b$
$\in$ $\Real$, such that $\hat{\mathbf{v}}=\R \hat{\mathbf{z}}$ is the direction
of the line and $\mathbf{d}=\R(a \hat{\mathbf{x}} + b\hat{\mathbf{y}})$ is the
closest point between the line and the origin. A line $\Line$ can therefore be
defined as:
\begin{equation}
\Line = \left\{\langle \R,(a, b)\rangle \in \SOthree \times
\mathbb{R}^{2}\right\}
\end{equation}
Let $\boxtimes$ be the operator that applies a transform $\T_{ij} = (\R_{ij},
\tran_{ij})$ to all the points of a line $\Line_i$ to get $\Line_j$ such that:
\begin{align}
 \R_j &= \R_{ij} \R_i \nonumber\\
 a_j &= a_i -
\left[\begin{smallmatrix} 1 & 0 & 0 \end{smallmatrix}\right] 
\R^\transpose_{ij} \tran_{ij} \nonumber\\
b_j &= b_i - \left[\begin{smallmatrix} 0 & 1 & 0 \end{smallmatrix}\right] 
\R^\transpose_{ij} \tran_{ij}
\label{eq:line-transform}
\end{align}
The error operator $\boxminus$ between two lines
$\Line_i,\Line_j$ is defined as:
\begin{equation}
	\Line_i \boxminus \Line_j = \left(\left[\begin{smallmatrix}
	                                     1 &0 \\
	                                     0& 1\\
                                         0& 0
	                                  \end{smallmatrix}\right]^\transpose
\Log{\R_i^\transpose \R_j}, a_i - a_j, b_i - b_j
	\right) \in \mathbb{R}^{4}
	\label{eq:line-minus}
\end{equation}
Given \Equation \ref{eq:line-transform} and \Equation \ref{eq:line-minus}, the
residual between a measured line $\tilde{\Line}_i$ and its prediction is defined
as follows:
\begin{equation}
	\mathbf{r}_{x_i, \Line\ell} = \left(\T^{-1}_i \boxtimes \Line_\ell\right) 
	\boxminus
\tilde{\Line}_i \label{eq:line-residual}
\end{equation}
We use the numerical derivatives of \Equation (\ref{eq:plane-residual}) and
(\ref{eq:line-residual}) in the optimization, using the symmetric difference
method.

\section{Implementation}
\label{sec:implementation}

\begin{figure}
	\vspace{2mm}
	\centering
	\includegraphics[width=0.85\columnwidth]{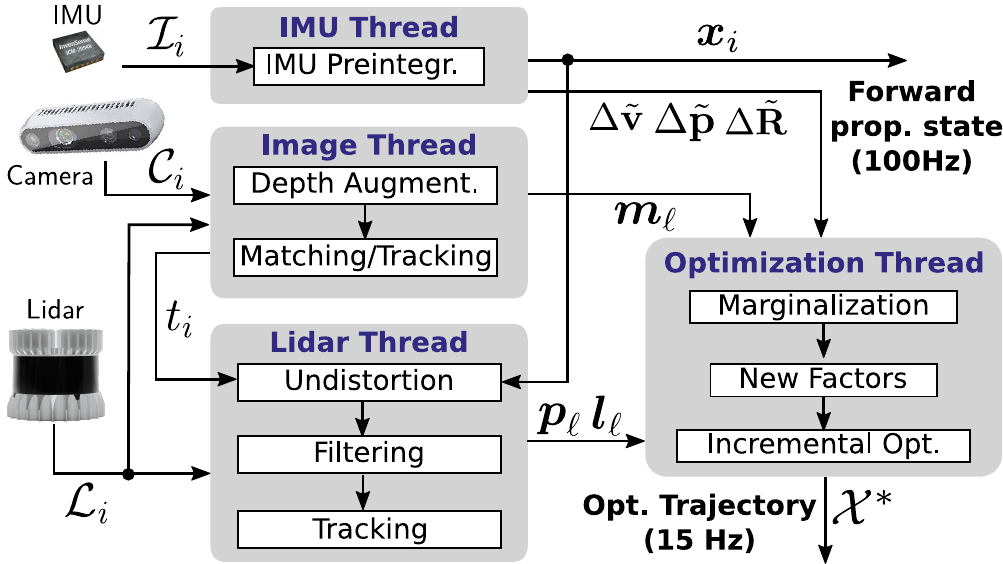}
	\caption{Overview of the VILENS system architecture. The inputs are handled
	in separate threads by each front-end measurement handler. The back-end
	produces both a high frequency forward-propagated output and a lower
	frequency optimized output. This parallel architecture allows for different
	measurement inputs depending on the platform.}
	\label{fig:state-estimation-architecture}
	\vspace{-3mm}
\end{figure}

The system architecture is shown in \Figure
\ref{fig:state-estimation-architecture}. Using four parallel threads for the
sensor processing and optimization, the system outputs the state estimated by
the factor-graph at camera keyframe frequency (typically \SI{15}{\hertz}) and
the IMU forward-propagated state at IMU frequency (typically \SI{100}{\hertz})
for use in navigation/mapping and control respectively.

The factor graph is solved using a fixed lag smoothing framework based on the
efficient incremental optimization solver iSAM2, using the GTSAM library
\cite{Dellaert2017}. For these experiments, we use a lag time of between \num{5}
and \SI{10}{\second}. All visual and lidar factors are added to the graph using
the Dynamic Covariance Scaling (DCS) \cite{MacTavish2015} robust cost function
to reduce the effect of outliers.

\subsection{Visual Feature Tracking}
We detect features using the FAST corner detector, and track them between
successive frames using the KLT feature tracker with outliers rejected using
RANSAC. Thanks to the parallel architecture and incremental optimization, every
second frame is used as a keyframe, achieving \SI{15}{\hertz} nominal output.

\subsection{Lidar Processing and Feature Tracking}
A key feature of our algorithm is that we extract feature primitives from the
lidar point clouds represented at the same time as a camera frame, such that the
optimization can be executed for all the sensors at once. The processing
pipeline consists of the following steps: point cloud undistortion and
synchronization, filtering, primitive extraction and tracking, and factor
creation.

\subsubsection{Undistortion and Synchronization}
\label{sec:lidar-sync}
\begin{figure}
 \vspace{2mm}
 \centering
 \includegraphics[width=0.9\columnwidth]{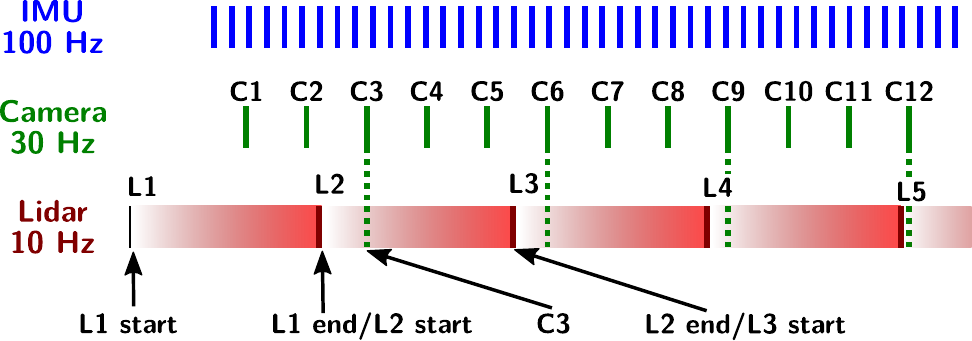}
 \caption{Example of output frequencies and synchronization between IMU, camera
and lidar. IMU and camera signals are captured instantaneously, while lidar
points are captured and accumulated for \SI{0.1}{\second} before being sent as a
scan. When the scan L2 is received, it is undistorted to the same time as camera
frame C3 and added to the same graph node as the camera.}
\label{fig:timestamps}
\vspace{-3mm}
\end{figure}

\Figure \ref{fig:timestamps} compares the different output frequencies of our
sensors. While IMU and camera samples are captured instantaneously, lidars
continually capture points while their internal mirror rotates around the
$z$-axis. Once a full rotation is complete, the accumulated laser returns are
converted into a point cloud and a new scan starts immediately thereafter.

Since the laser returns are captured while moving, the point cloud needs to be
undistorted with a motion prior and associated to a unique arbitrary timestamp
-- typically the start of the scan \cite{mc2slam}. This approach would imply
that camera and lidar measurements have different timestamps and thus separate
graph nodes.

Instead, we choose to undistort the lidar measurement to the closest camera
timestamp after the start of the scan. For example, in \Figure
\ref{fig:timestamps}, the scan L2 is undistorted to the timestamp of keyframe
C3. Given the forward propagated states from the IMU module, the motion prior is
linearly extrapolated using the timestamp associated to each point of the cloud
(for simplicity, we avoid Gaussian-Process interpolation \cite{le2020in2laama}
or state augmentation with time offsets \cite{zebedee}). As the cloud is now
associated with C3, the lidar landmarks are connected to the same node as C3
rather than creating a new one.

This subtle detail not only guarantees that a consistent number of new nodes and
factors are added to the graph optimization, but it also ensures that the
optimization is performed \emph{jointly} between IMU, camera and lidar inputs.
This also ensures a fixed output frequency, \ie the camera framerate or lidar
framerate (when cameras are unavailable), but not a mixture of the two.

\subsubsection{Filtering}
Once the point cloud has been undistorted, we perform the segmentation from
\cite{Bogoslavskyi2016} to separate the points into clusters. Small clusters
(less than 5 points) are marked as outliers and discarded as they are likely to
be noisy.

Then, the local curvature of each point in the pre-filtered cloud is calculated
using the approach of \cite{shan-legoloam}. The points with the lowest and
highest curvature are assigned to the set of plane candidates $\calC_P$ and line
candidates $\calC_L$, respectively.

The segmentation and curvature-based filtering typically reduce the number of
points in the point cloud by 90\%, providing significant computational savings
in the subsequent plane and line processing.

\subsubsection{Plane and Line Extraction and Tracking}
Over time, we track planes and lines in the respective candidate sets $\calC_P$
and $\calC_L$. This is done in a manner analogous to local visual feature
tracking methods, where features are tracked within a local vicinity of their
predicted location.

First, we take the tracked planes and lines from the previous scan,
$\Plane_{i-1}$ and $\Line_{i-1}$,  and use the IMU forward propagation to
predict their location in the current scan, $\hat{\Plane}_i$ and
$\hat{\Line}_i$. Then to assist local tracking, we segment $\calC_P$ and
$\calC_L$ around the predicted feature locations using a maximum point-to-model
distance. Afterwards, we perform Euclidean clustering (and normal filtering for
plane features) to remove outliers. Then, we fit the model to the segmented
point cloud using a PROSAC \cite{Chum2005-prosac} robust fitting algorithm.

Finally, we check that the predicted and detected landmarks are sufficiently
similar. Two planes, $\Plane_i$ and $\Plane_j$, are considered a match when
difference between their normals and the distance from the origin are smaller
than a threshold:
\begin{eqnarray}
\delta_{n} = & \left\Vert \arccos(\hat{\mathbf{n}}_i \cdot
\hat{\mathbf{n}}_j) \right\Vert
& < \alpha_p \\
\delta_{d} = & \left\Vert \hat{\mathbf{n}}_i d_i -
\hat{\mathbf{n}}_j d_j \right\Vert & < \beta_p
\end{eqnarray}

\noindent Two lines $\Line_i$ and $\Line_j$ are considered a match if their
directions and their center distances are smaller than a threshold:
\begin{eqnarray}
	\delta_{n} = & \left\Vert \arccos(\hat{\mathbf{v}}_i \cdot
	\hat{\mathbf{v}}_j)
	\right\Vert & < \alpha_l \\
	\delta_{d} = &  \left\Vert (\mathbf{d}_i - \mathbf{d}_j) - 
	\left((\mathbf{d}_i -
	\mathbf{d}_j)\cdot
	\hat{\mathbf{v}}_i \right) \hat{\mathbf{v}}_i \right\Vert  & < \beta_l
\end{eqnarray}
In our case $\alpha_p = \alpha_l = 0.35~\textnormal{rad},~\beta_p = \beta_l =
0.5~\textnormal{m}$.

Once a feature has been tracked, the feature's inliers are removed from the
corresponding candidate set, and the process is repeated for the remaining
landmarks.

After tracking is complete, we detect new landmarks in the remaining candidate
clouds. The point cloud is first divided using Euclidean clustering for lines,
and normal-based region growing for planes. We then detect new landmarks in each
cluster using the same method as landmark tracking.

Point cloud features are only included in the optimization after they have been
tracked for a minimum number of consecutive scans. Note that the oldest features
are tracked first, to ensure the longest possible feature tracks.

\subsection{Zero Velocity Update Factors}
To limit drift and factor graph growth when the platform is stationary, we add
zero velocity constraints to the graph when updates from two out of three
modalities (camera, lidar, IMU) report no motion.

\section{Experimental Results}
\label{sec:results}

\begin{figure}
\vspace{2mm}
\centering
\includegraphics[width=0.42\columnwidth]{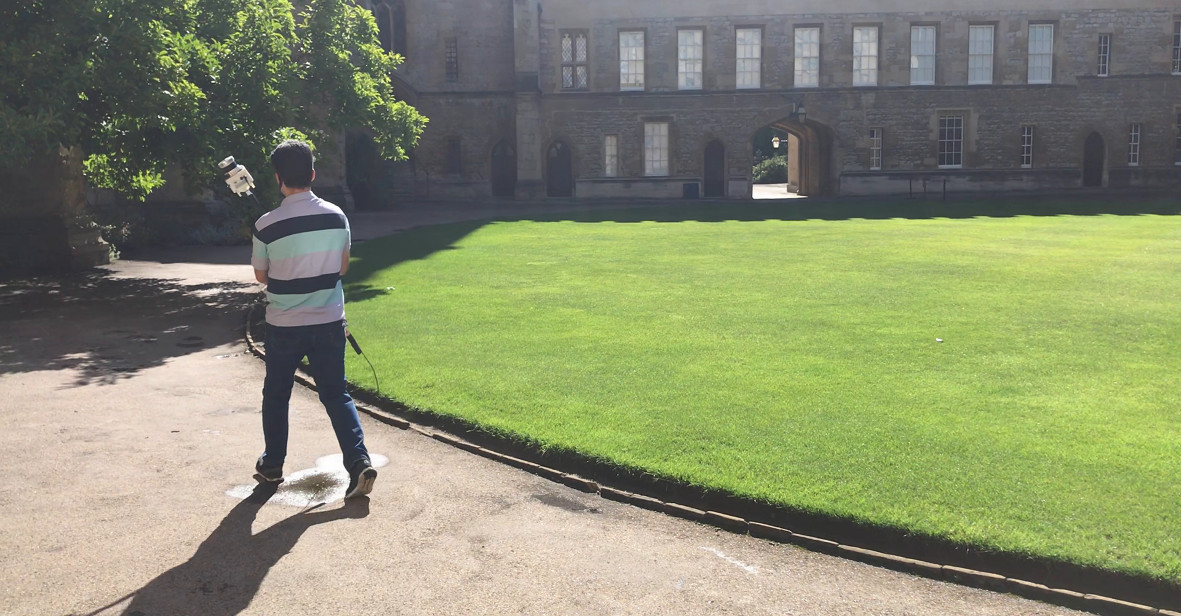}%
 \hspace{1mm}%
\includegraphics[width=0.42\columnwidth]{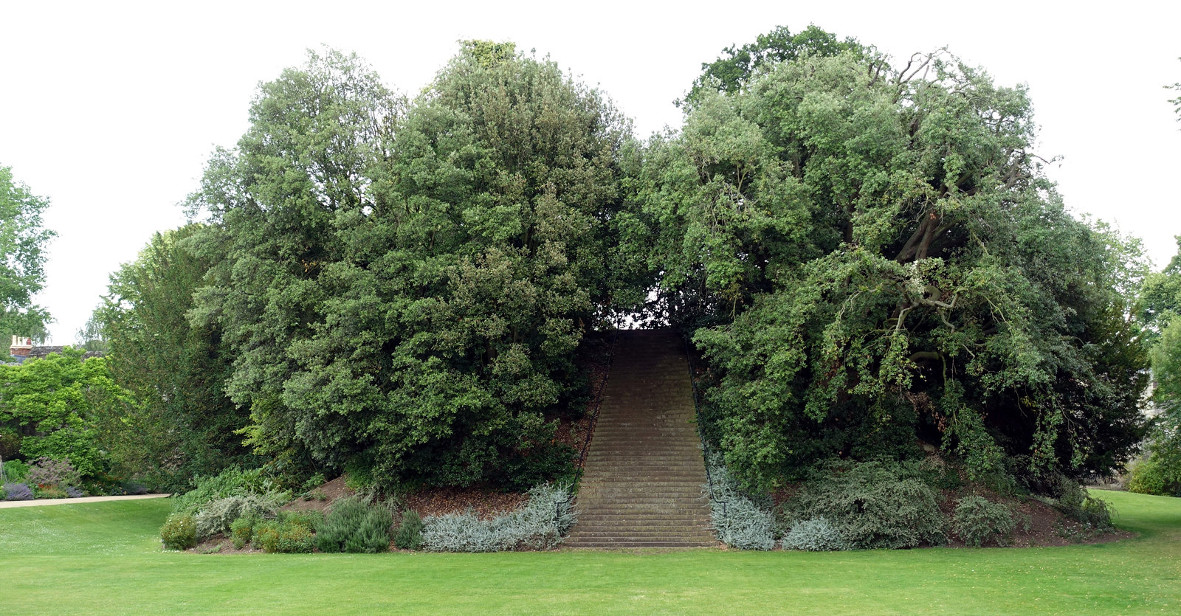}\\
 \vspace{1mm}%
\includegraphics[width=0.42\columnwidth]{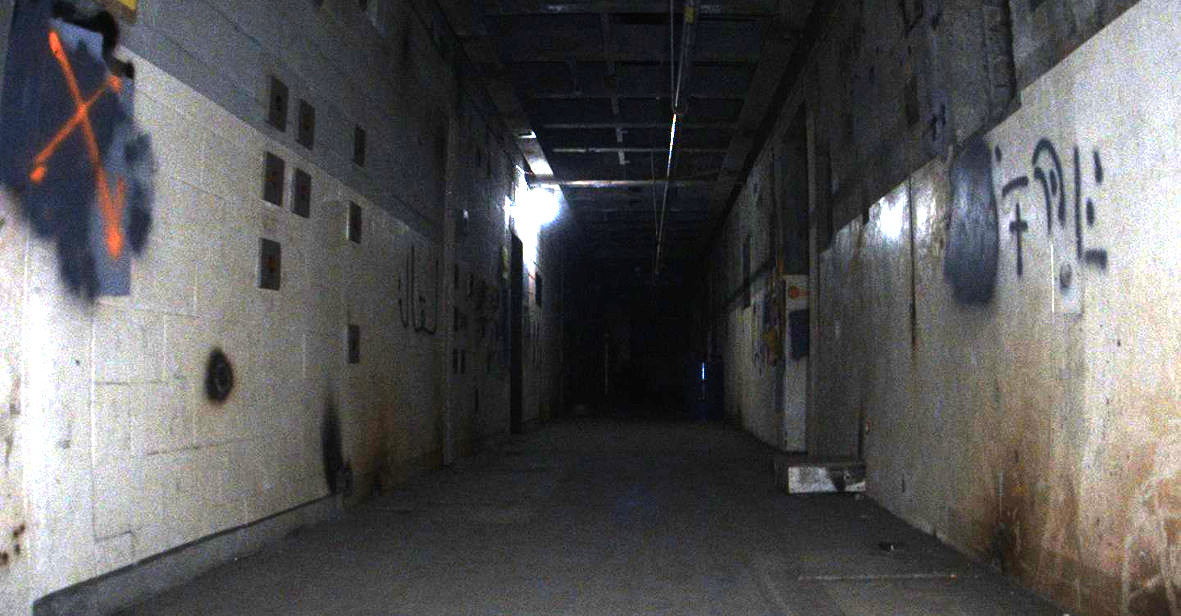}%
 \hspace{1mm}%
\includegraphics[width=0.42\columnwidth]{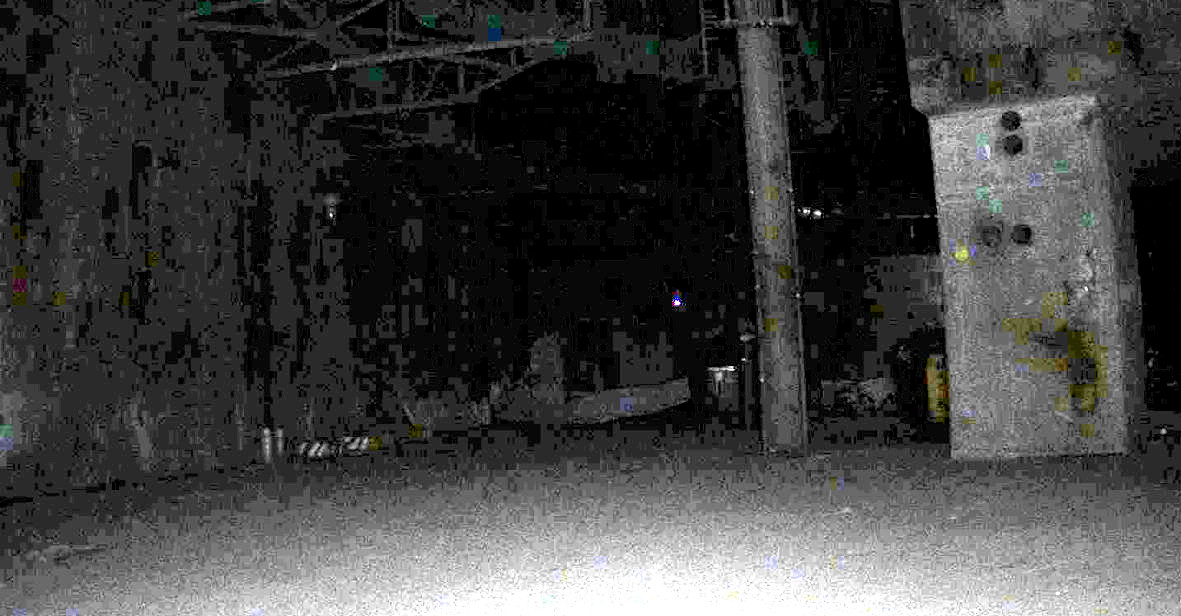}
\caption{\textit{Top:} The Newer College dataset \cite{ramezani2020newer} --
challenges include large open spaces, dense foliage without clear structure, and
large illumination changes from sunlight to shadow. \textit{Bottom:} The DARPA
SubT dataset -- challenges include long straight corridors and low light
conditions (the bottom images were manually enhanced to show the content).}
\label{fig:experimental-scenarios}
\vspace{-2mm}
\end{figure}

We evaluated our algorithm on a variety of indoor and outdoor environments in
two contrasting datasets: the Newer College Dataset \cite{ramezani2020newer} and
the DARPA SubT Challenge (Urban). An overview of these environments is shown in
\Figure \ref{fig:experimental-scenarios}.

\subsection{Datasets}
The Newer College dataset (NC) \cite{ramezani2020newer} was collected using a
portable device equipped with a Ouster OS1-64 Gen1 lidar sensor, a RealSense
D435i stereo IR camera, and an Intel NUC PC. The cellphone-grade IMU embedded in
the lidar was used for inertial measurements. The device was carried by a person
walking outdoor surrounded by buildings, large open spaces, and dense foliage.
The dataset includes challenging sequences where the device was shaken
aggressively to test the limits of tracking.

The SubT dataset (ST) consists of two of the most significant runs of the SubT
competition (Alpha-2 and Beta-2) collected on two copies of the ANYmal B300
quadruped robot \cite{Hutter2016} equipped with a Flir BFS-U3-16S2C-CS monocular
camera and a industrial-grade Xsens MTi-100 IMU, which were hardware
synchronized by a dedicated board \cite{Tschopp2020}. A Velodyne VLP-16 was also
available but was synchronized via software. The robots navigated the
underground interiors of an unfinished nuclear reactor. This dataset is
challenging due to the presence of long straight corridors and extremely dark
environments. Note that the leg kinematics from the robot \textbf{was not used}
in this work.

The specific experiments are named as follows:
\begin{itemize}
\item \textbf{NC-1:} Walking around an open college environment
(\SI{1134}{\meter}, \SI{17}{\minute}).
\item \textbf{NC-2:} Walking with highly dynamic motion in the presence of
strong illumination changes (\SI{480}{\meter}, \SI{6}{\minute}).
\item \textbf{NC-3:} Shaking the sensor rig at very high angular velocities, up
to \SI{3.37}{\radian/\second} (\SI{91}{\meter}, \SI{2}{\minute}).
\item \textbf{ST-A:} Anymal quadruped robot trotting in dark underground reactor
facility (\SI{167}{\meter}, \SI{11}{\minute}).
\item \textbf{ST-B:} A different Anymal robot in a part of the reactor
containing a long straight corridor (\SI{490}{\meter}, \SI{60}{\minute}).
\end{itemize}

To generate ground truth, ICP was used to align the current lidar scan to
detailed prior maps, collected using a commercial laser mapping system. For an
in-depth discussion on ground truth generation the reader is referred to
\cite{ramezani2020newer}.

\begin{table}
  \vspace{3mm}
  \centering
  \begin{tabular}{l|ccc}  \toprule
  \multicolumn{4}{c}{\textbf{Relative Pose Error (RPE) -- Translation
  [\si{\metre}] / Rotation [\si{\degree}]}} \\
 \midrule
\multicolumn{4}{c}{\textit{SubT Challenge Urban (ANYmal B300)}} \\ \midrule
\textbf{Data} & \textbf{VILENS-LI}& \textbf{VILENS-LVI}& \textbf{LOAM}\\
\midrule
ST-A & 0.28 / 2.65 & \textbf{0.14} / \textbf{1.04} & 0.20 / 1.32\\
ST-B & 0.20 / 1.75 & \textbf{0.12} / \textbf{0.79} & 0.22 / 0.99\\
\midrule
\multicolumn{4}{c}{\textit{Newer College (Handheld %
		Device \cite{ramezani2020newer})}} \\ \midrule
\textbf{Data} & \textbf{VILENS-LI}& \textbf{VILENS-LVI}&
\textbf{LeGO-LOAM}\\
\midrule
NC-1 & 0.50 / 3.23 & 0.39 / \textbf{2.45} & \textbf{0.34} / 3.45 \\
NC-2 & 0.78 / 2.38 & \textbf{0.43} / \textbf{1.85} & 12.32 / 55.00 \\
NC-3 & \textbf{0.18} / \textbf{3.12} & 0.20 / 3.61 & 1.85 / 23.67 \\
\bottomrule
\end{tabular}
\vspace{1mm} %
\caption{Experimental Results}
\label{tab:rpe}
\vspace{-6mm}
\end{table}

\subsection{Results}
Table \ref{tab:rpe} summarizes the mean Relative Pose Error (RPE) over a
distance of \SI{10}{\meters} for the following algorithms:
\begin{itemize}
	\item \textbf{VILENS-LI:} VILENS with IMU and lidar only;
	\item \textbf{VILENS-LVI:} VILENS with IMU, visual (with lidar depth),
and lidar features;
	\item \textbf{LOAM:} The output of the LOAM \cite{Zhang_2014} mapping
	module used during the SubT competition.
	\item \textbf{LeGO-LOAM:} The output of the LeGO-LOAM \cite{shan-legoloam}
	mapping module.
\end{itemize}

It should be noted that no loop closures have been performed, and in contrast
to both LOAM and LeGO-LOAM methods we do not perform any mapping.

\begin{figure}[!t]
\vspace{3mm}
 \centering
 \includegraphics[width=1\columnwidth]{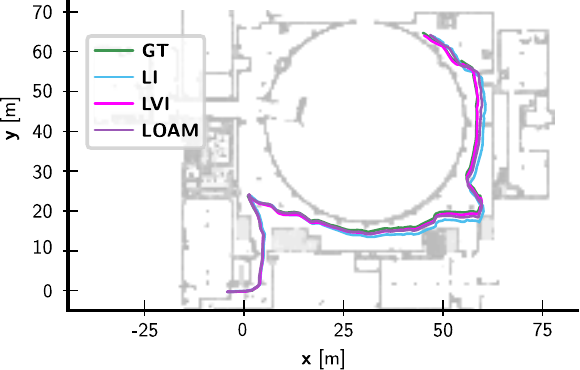}\\
 \includegraphics[width=1\columnwidth]{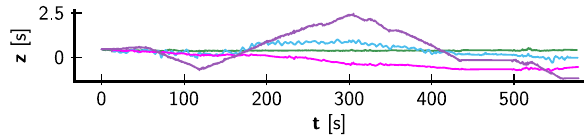}\\
 \caption{Aerial view and elevation over time on the
\textbf{ST-A} dataset, showing the estimated
trajectory with Lidar-Inertial (blue), Lidar-Visual-Inertial (magenta), and
LOAM \cite{Zhang_2014} against the ground truth (green). Note that there are no
loop-closures present in this system.}
\label{fig:subt-trajectory}
\vspace{-3mm}
\end{figure}

For the SubT datasets, VILENS-LVI outperforms LOAM in translation / rotation by
an average of 38\% / 21\% and VILENS-LI by 46\% / 21\%. An example of the global
performance is shown in \Figure \ref{fig:subt-trajectory}, which depicts both
the estimated and ground truth trajectories on the ST-A dataset. VILENS-LVI is
able to achieve very slow drift rates, even without a mapping system or loop
closures.

For the least dynamic NC dataset, NC-1, VILENS-LVI achieves comparable
performance to LeGO-LOAM. However, For the more dynamic datasets
(up to \SI{3.37}{\radian/\second}), NC-2 and NC-3, the VILENS methods
significantly outperform
LeGO-LOAM. Key to this performance is the undistortion of the
lidar cloud to the camera timestamp, allowing accurate visual feature
depth-from-lidar, while minimizing computation.

Overall, the best performing algorithm was VILENS-LVI, showing how the tight
integration of visual and lidar features allows us to avoid failure modes that
may be present in lidar-inertial only methods.

\subsection{Multi-Sensor Fusion}
A key benefit arising from the tight fusion of complementary sensor modalities
is a natural robustness to sensor degradation. While a significant proportion of
the datasets presented favorable conditions for both lidar and visual feature
tracking, there were a number of scenarios where the tight fusion enabled
increased robustness to failure modes of individual sensors.

\begin{figure}
\vspace{3mm}
\centering
\includegraphics[width=0.43\columnwidth]{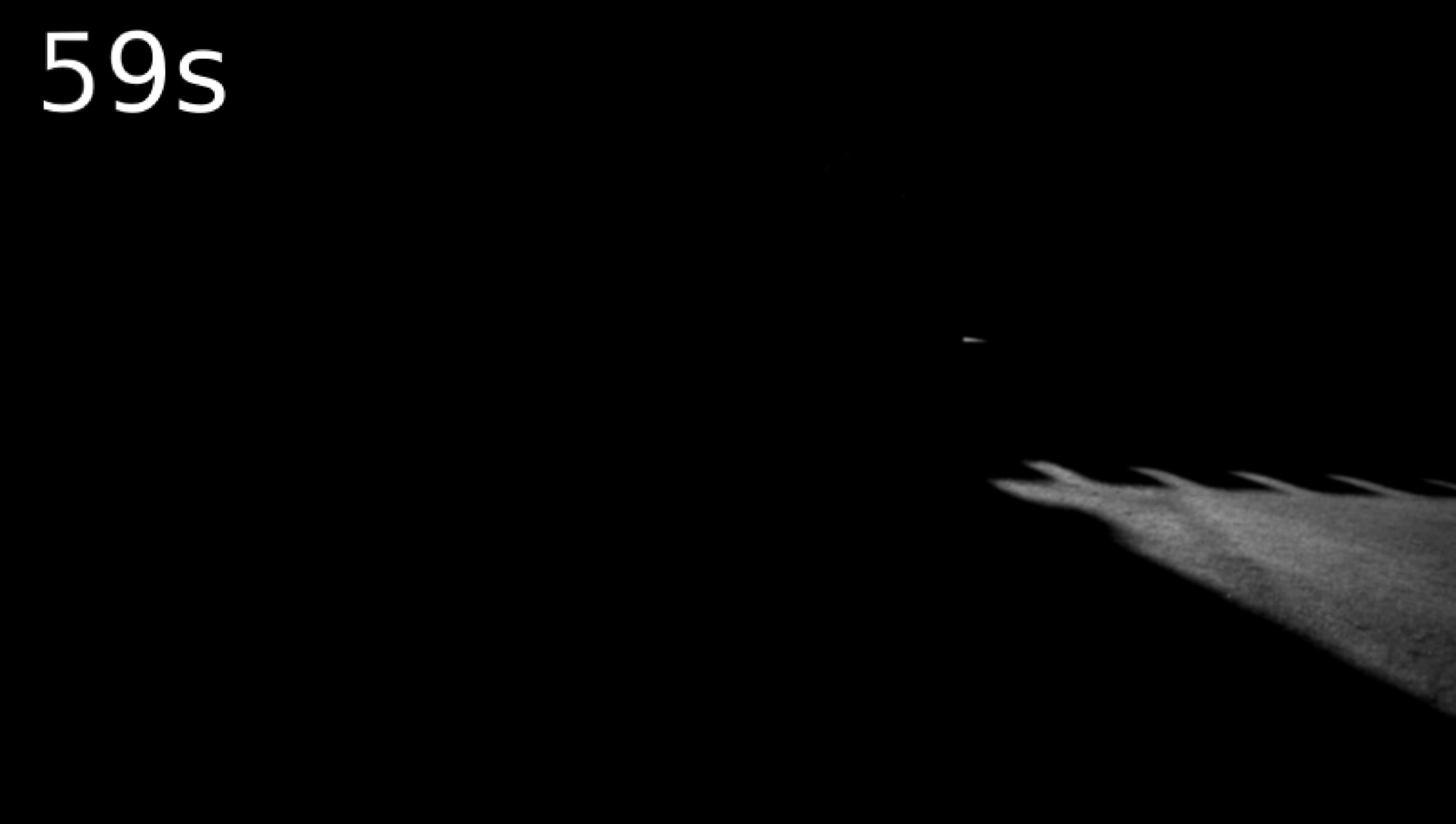}
\hspace{0.01\columnwidth}%
\includegraphics[width=0.43\columnwidth]{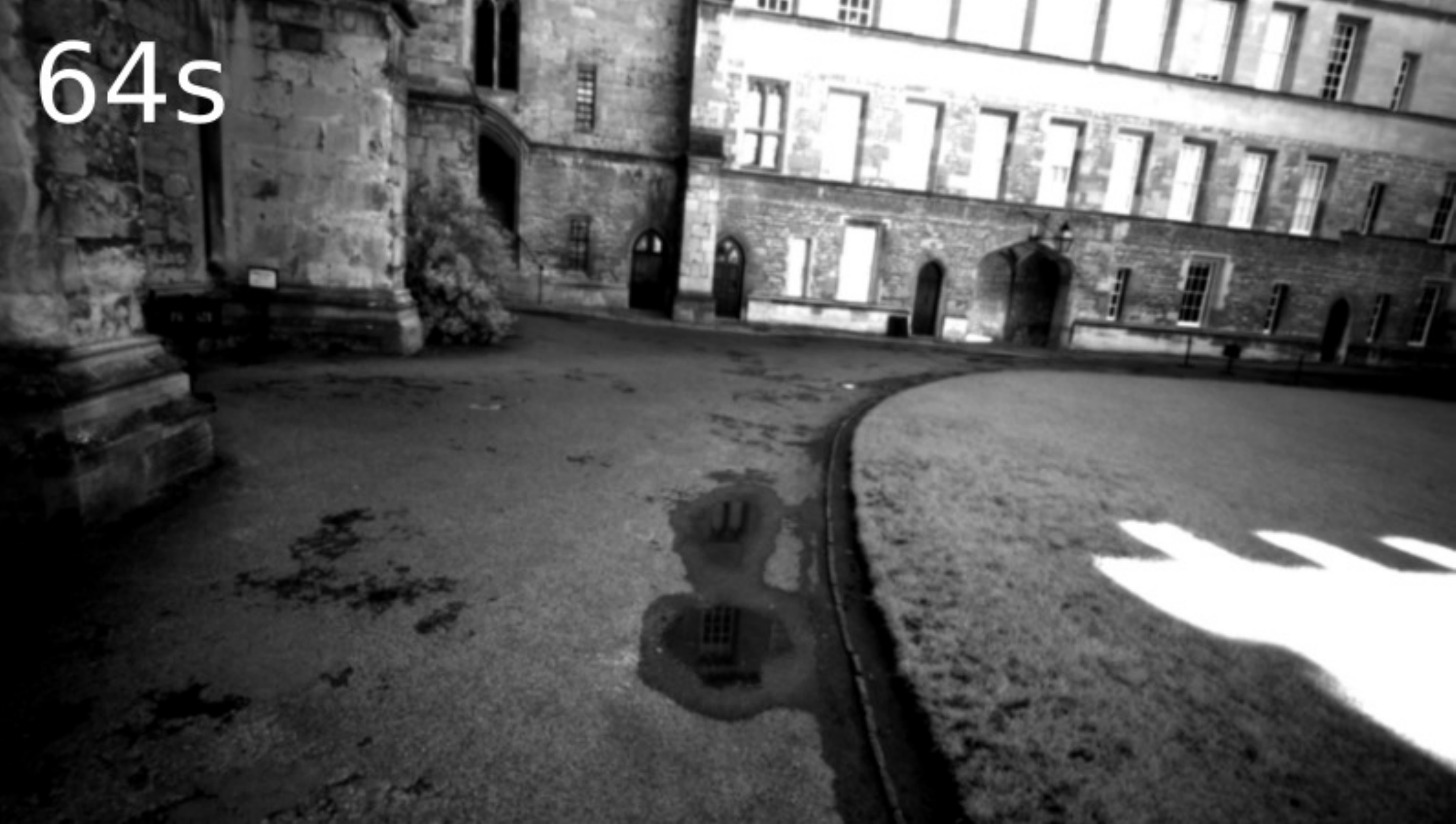}%
\\
\vspace{0.01\columnwidth}
\includegraphics[width=\columnwidth]{%
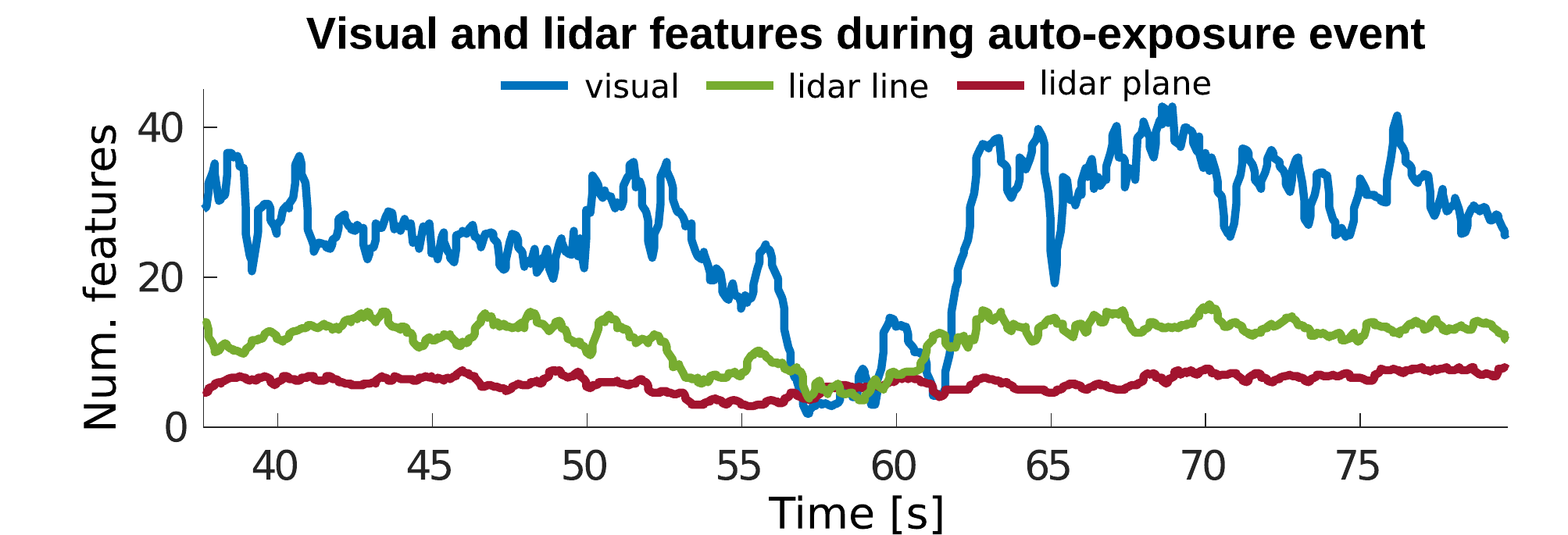}
\caption{\textit{Top:} The Newer College (\textbf{NC-2}) contains sections with
dramatic exposure change from underexposure (left) to more balanced exposure
(right). \textit{Bottom:} During this auto-exposure adjustment (\SI{57}{\second}
to \SI{62}{\second}) the number of visual features (blue) decreases almost to
zero, while the lidar plane (red) and line (green) feature count remains
relatively constant.}
\label{fig:visual-feature-dropout}
\end{figure}

\Figure \ref{fig:visual-feature-dropout} shows an example from the NC-2 where
the camera auto-exposure feature took \SI{\sim 3}{\second} to adjust when moving
out of bright sunlight into shade. During this time the number of visual
features drops from around 30 to less than 5 (all clustered in a corner of the
image). This would cause instability in the estimator. By tightly fusing the
lidar, we are able to use the small number of visual features and the lidar
features, without causing any degradation in performance. This is in contrast to
methods such as \cite{pronto,Khattak2020} where the use of separate
visual-inertial and lidar-inertial subsystems mean that degenerate situations
must be explicitly handled.

Similarly, in cases where the lidar landmarks are not sufficient to fully
constrain the estimate (or are close to degenerate), the tight fusion of visual
features allow the optimisation to take advantage of the lidar constraints while
avoiding problems with degeneracy.

\subsection{Analysis}
\label{sec:analysis}

A key benefit from using light-weight point cloud primitives in the optimisation
is improved efficiency. The mean computation times for the above datasets are
\SI{\sim 10}{\milli\second} for visual feature tracking, \SI{\sim
50}{\milli\second} for point cloud feature tracking, and \SI{\sim
20}{\milli\second} for optimization on a consumer grade laptop. This enables the
system to output at \SI{10}{\hertz} (lidar frame rate) when using lidar-inertial
only, and \SI{15}{\hertz} (camera keyframe rate) when  fusing vision, lidar, and
inertial measurements.

\section{Conclusion}
\label{sec:conclusions}
We have presented a novel factor graph formulation for state estimation that
tightly fuses camera, lidar, and IMU measurements. This fusion enables for
graceful handling of degenerate modes -- blending between lidar-only feature
tracking and visual tracking (with lidar depth), depending on the constraints
which each modality can provide in a particular environment. We have
demonstrated comparable performance to state-of-the-art lidar-inertial odometry
systems in typical conditions and better performance in extreme conditions, such
as aggressive motions or abrupt light changes. Our approach also presents a
novel method of jointly optimizing lidar and visual features in the same factor
graph. This allows for robust estimation in difficult environments such as long
corridors, and dark environments.

\bibliographystyle{./IEEEtran}
\bibliography{./IEEEabrv,./library}

\end{document}